\documentclass{article}

\usepackage{arxiv}

\usepackage[utf8]{inputenc} 
\usepackage[T1]{fontenc}    
\usepackage{hyperref}       
\usepackage{url}            
\usepackage{booktabs}       
\usepackage{amsfonts}       
\usepackage{nicefrac}       
\usepackage{microtype}      
\usepackage{graphicx}
\usepackage{natbib}
\usepackage{doi}
\usepackage{amsmath}

\title{Neural Attention: Enhancing QKV Calculation in Self-Attention Mechanism with Neural Networks}

\author{
  Muhan Zhang \\
  Fuzhou University \\
  \texttt{wddzmh@gmail.com} \\
}

\date{}
\renewcommand{\shorttitle}{Enhancing QKV Calculation in Self-Attention}

\hypersetup{
pdftitle={Neural Attention: Enhancing QKV Calculation in Self-Attention Mechanism with Neural Networks},
pdfsubject={Deep Learning, Self-Attention, Neural Networks},
pdfauthor={Muhan Zhang},
pdfkeywords={Self-Attention, QKV Calculation, Neural Networks, Deep Learning},
}

\begin{document}
\maketitle

\begin{abstract}
	In the realm of deep learning, the self-attention mechanism has substantiated its pivotal role across a myriad of tasks, encompassing natural language processing and computer vision. Despite achieving success across diverse applications, the traditional self-attention mechanism primarily leverages linear transformations for the computation of query, key, and value (QKV), which may not invariably be the optimal choice under specific circumstances. This paper probes into a novel methodology for QKV computation—implementing a specially-designed neural network structure for the calculation. Utilizing a modified Marian model, we conducted experiments on the IWSLT 2017 German-English translation task dataset and juxtaposed our method with the conventional approach. The experimental results unveil a significant enhancement in BLEU scores with our method. Furthermore, our approach also manifested superiority when training the Roberta model with the Wikitext-103 dataset, reflecting a notable reduction in model perplexity compared to its original counterpart. These experimental outcomes not only validate the efficacy of our method but also reveal the immense potential in optimizing the self-attention mechanism through neural network-based QKV computation, paving the way for future research and practical applications. The source code and implementation details for our proposed method can be accessed at \url{https://github.com/ocislyjrti/NeuralAttention}.

\end{abstract}


\section{Introduction}

\subsection{Problem Statement and Research Motivation}

The self-attention mechanism, introduced by Vaswani et al. in their seminal work on the Transformer architecture~\cite{vaswani2017attention}, has established itself as a potent method in capturing dependencies within input sequences across various tasks in natural language processing and computer vision. Despite its notable achievements, the traditional self-attention mechanism, which computes the query, key, and value (QKV) predominantly through linear transformations, might encounter limitations in its expressive power in certain scenarios. As highlighted by Goodfellow et al. in their comprehensive book on deep learning~\cite{goodfellow2016deep}, linear transformations essentially perform linear mapping, potentially lacking the capability to handle complex patterns and non-linear relationships. In contrast, non-linear transformations, such as those performed through neural networks, generally possess the ability to capture more intricate features and patterns in input data. Thus, the main motivation of this paper is to explore a novel method, leveraging neural networks to enhance QKV computation within the self-attention mechanism, aiming to amplify its expressive power and performance.

\subsection{Research Significance}

Such an enhancement in the self-attention mechanism is profoundly significant. As shown by its application in machine translation with the Transformer architecture~\cite{vaswani2017attention} and in image recognition with the Vision Transformer (ViT)~\cite{dosovitskiy2020image}, any improvement upon this mechanism may directly impact a multitude of applications and models dependent on it.

\subsection{Proposed Solution and Contributions}

Against this backdrop, we propose a novel approach to compute QKV in the self-attention mechanism using a neural network, and validate its efficacy through experiments on different datasets and models. Our main contributions are as follows:

\begin{enumerate}
    \item Proposing and implementing a novel neural network model for computing QKV in the self-attention mechanism, elucidating its architecture and underlying rationale, inspired by recent advancements in neural attention~\cite{lin2017structured}.
    \item Validating the effectiveness of our method in enhancing model performance through experiments using a modified Marian model on the IWSLT 2017 German-English translation task dataset and a modified Roberta model~\cite{liu2019roberta} on the Wikitext-103 dataset.
\end{enumerate}

\section{Background}

\subsection{The Rise of the Self-Attention Mechanism}
Since its groundbreaking introduction by Vaswani et al. within the Transformer model, the self-attention mechanism has rapidly grown in prominence and influence, becoming a cornerstone in the design of many subsequent models and architectures in the expansive field of deep learning. This ingenious mechanism, with its unique ability to process input sequences concurrently in parallel, has redefined how models perceive and handle dependencies. Unlike traditional models that often struggled with long-range dependencies, the self-attention mechanism excels by capturing intricate relationships between elements, irrespective of the vast positional distances that might separate them in the input sequence. Such capabilities not only enhance the model's understanding of the data but also ensure more contextually accurate representations. Over time, this mechanism has proven its mettle, showing remarkable performance improvements across a diverse range of tasks, from machine translation to document summarization, solidifying its position as an indispensable tool in modern deep learning toolkits.

\subsection{Roberta Model}
The Roberta model, short for "Robustly Optimized BERT Approach", represents an evolution of the BERT architecture~\cite{devlin2018bert}. While the foundational principles remain similar, Roberta introduces significant optimizations, particularly in terms of model size and the approach to pre-training data. One of the strategic changes was the decision to eliminate the "Next Sentence Prediction" task from BERT, and instead, Roberta adopts longer training sequences. This design choice has proven to be pivotal, resulting in increased efficiency and precision. Consequently, Roberta has showcased superior performance and robust representational capacities across a broad spectrum of natural language processing tasks~\cite{liu2019roberta}. The underlying design philosophies and optimization strategies have cemented Roberta's position as an invaluable asset in the NLP research and application landscape, garnering acclaim from many researchers and practitioners.

\subsection{Marian Model}
Marian NMT is an efficient, free, and open-source framework specifically designed for research in neural machine translation (NMT)~\cite{junczys2018marian}. Supporting multi-GPU training and model ensemble, it provides implementations of a range of model architectures and training strategies, including the standard Transformer model and various derivatives. Marian has achieved success across various machine translation benchmark tests.

In the following sections, we will delve into the method proposed, which involves utilizing neural networks to enhance QKV computation within the self-attention mechanism, and observe the results from experiments based on Roberta and Marian models.

\section{Proposed Method and Theoretical Analysis}

\subsection{Method Overview}

In this section, we present an enhanced attention mechanism, which introduces a Multilayer Perceptron (MLP) to augment the model's representational learning capability \cite{rumelhart1986learning}. Our motivation stems from the notion that traditional linear attention mechanisms, such as those introduced by Vaswani et al. \cite{vaswani2017attention}, may not sufficiently capture complex patterns and non-linear relationships within the input data.

\subsection{Detailed Methodology}

The core computation in traditional attention mechanisms is typically represented as

\[
\text{Attention}(Q, K, V) = \text{softmax}\left( \frac{{QK^\top}}{{\sqrt{d_k}}} \right) V
\]

where \(Q\), \(K\), and \(V\) are commonly obtained through a single linear transformation:

\[
Q = W_qX, \quad K = W_kX, \quad V = W_vX
\]

In the method we propose, we employ an MLP with a certain depth to replace these linear transformations, formally we have:

\[
Q = \text{MLP}_q(X), \quad K = \text{MLP}_k(X), \quad V = \text{MLP}_v(X)
\]

Where the MLP can be expressed as:

\[
\text{MLP}(X) = W_2 \cdot \sigma\left(\text{LayerNorm}(W_1X + b_1)\right) + b_2
\]

Where:
\begin{itemize}
    \item \(X\) is the input,
    \item \(W_1\) and \(b_1\) are the weight and bias of the first layer, respectively,
    \item \(\sigma\) represents the ReLU activation function \cite{nair2010rectified},
    \item \(\text{LayerNorm}\) denotes the Layer Normalization operation \cite{ba2016layer},
    \item \(W_2\) and \(b_2\) are the weight and bias of the second layer, respectively.
\end{itemize}

\subsection{Theoretical Analysis}

In the multi-head self-attention mechanism, each "head" learns different attention weights, thereby capturing different facets of information from the input sequence. The calculations for Query (Q), Key (K), and Value (V) play a pivotal role in this process. In the original attention mechanism, these vectors are obtained through linear transformations. These transformations can be understood as operations that reposition word vectors within the vector space.

From a geometric or vector space perspective, linear transformations include operations such as rotation, scaling, and translation \cite{goodfellow2016deep}. Therefore, the traditional linear attention mechanism moves word vectors in space to focus on relevant context. Specifically, by learning linear transformations for Q, K and V, the model learns how to reposition word vectors in space so that vectors of words that are associated or need to be attended to are brought closer together.

\[
Q = W_qX, \quad K = W_kX, \quad V = W_vX
\]

However, a limitation of linear transformations is that they preserve the linear structure of the vector space. In other words, linear transformations cannot alter the topological structure of the space. In some cases, nonlinear relationships and complex patterns might not be sufficiently captured by linear transformations.

This is why we introduce the Multi-Layer Perceptron (MLP). Unlike single-layer linear transformations, an MLP, owing to its internal non-linear activation functions (e.g., ReLU), can implement nonlinear mappings from the input space to the output space \cite{rumelhart1986learning}. Therefore, the MLP has the capacity to alter the topological structure of the word vector space, potentially creating more complex and expressive relationships in the space.

\[
Q = \text{MLP}_q(X), \quad K = \text{MLP}_k(X), \quad V = \text{MLP}_v(X)
\]

In this way, our model can learn more complex relationships and dependencies between words, further enriching the semantic and structural information of the representations. Although this increases the computational complexity of the model, we argue that if this complexity brings about a significant improvement in performance, then the trade-off is worthwhile. In the subsequent experimental section, we will validate this through a series of experiments.

\section{Experiments}

\subsection{Experimental Setup}

Experiments were conducted based on the Hugging Face Transformers library \cite{wolf-etal-2020-transformers}. We assessed Roberta and Marian models on masked language modeling and machine translation tasks respectively. In the implementation, we employed pre-trained weights for model initialization, ensuring that parts of the model (excluding QKV computation) could benefit from the pre-trained weights, and ensured that both models used entirely identical parameters and settings during the experiments. All models were trained using the same default parameters, including the same learning rate, optimizer, batch size, etc. Furthermore, we ensured experiment consistency and reproducibility by setting the same random seed. These measures ensured that all observed performance differences can be attributed to variations in the QKV computation method, rather than other factors.

Firstly, we analyzed the machine translation task utilizing the opus-mt-de-en model pre-trained by the Helsinki-NLP team \cite{TiedemannThottingal:EAMT2020}, which is based on the MarianMTModel architecture, specifically designed for neural machine translation. Both the encoder and decoder of the model contain 6 layers, with hidden layer dimensions of 512, feed-forward layer dimensions of 2048, employing 8 attention heads, and applying a dropout rate of 0.1 in the model to suppress overfitting and enhance generalization capabilities. The source language (German) and target language (English) share a vocabulary of 58,101 words. We utilized the Hugging Face Transformers library to conduct further fine-tuning and evaluation of the model on the "IWSLT 2017" dataset (config name "iwslt2017-de-en") by executing the run\_translation.py script. The batch size for training and evaluation per device was set to 32, the model was trained on the data for 6 epochs, evaluated every 1000 training steps, and logged every 200 steps. All model outputs and logs were saved in the "/tmp/tst-translation" directory for further analysis and model inspection.

Secondly, we trained and evaluated the performance of the roberta-base model on the Masked Language Modeling (MLM) task. Specifically, the model adopts the RobertaForMaskedLM architecture, contains 12 encoder and decoder layers, with a hidden dimension of 768 per layer, feed-forward network layer dimensions of 3072, and deploys 12 attention heads. A dropout rate of 0.1 is used in the model, along with a vocabulary of 50,265 words. This experiment was conducted on the wikitext-103-raw-v1 dataset, utilizing the run\_mlm.py script for model fine-tuning and evaluation. In both training and evaluation stages, we set the batch size to 8, the model was trained on the data for 5 epochs, evaluated every 250 training steps, and logged every 50 steps. Relevant outputs and logs were saved in the /tmp/test-mlm directory. In both tasks, we particularly focused on the performance variations and results brought about by the fine-tuning process of the model on specific tasks and datasets.

\subsection{Experiment One: Roberta Model on Masked Language Modeling Task}

\subsubsection{Model Modification and Training Setup}

We modified the self-attention mechanism of the Roberta model, replacing the QKV computation from the original linear transformation to the MLP neural network. The model was trained and evaluated on the Wikitext-103-raw-v1 dataset, undergoing a total of 5 training epochs.

\subsubsection{Results}

\begin{figure}[htbp]
\centering
\includegraphics[width=0.7\textwidth]{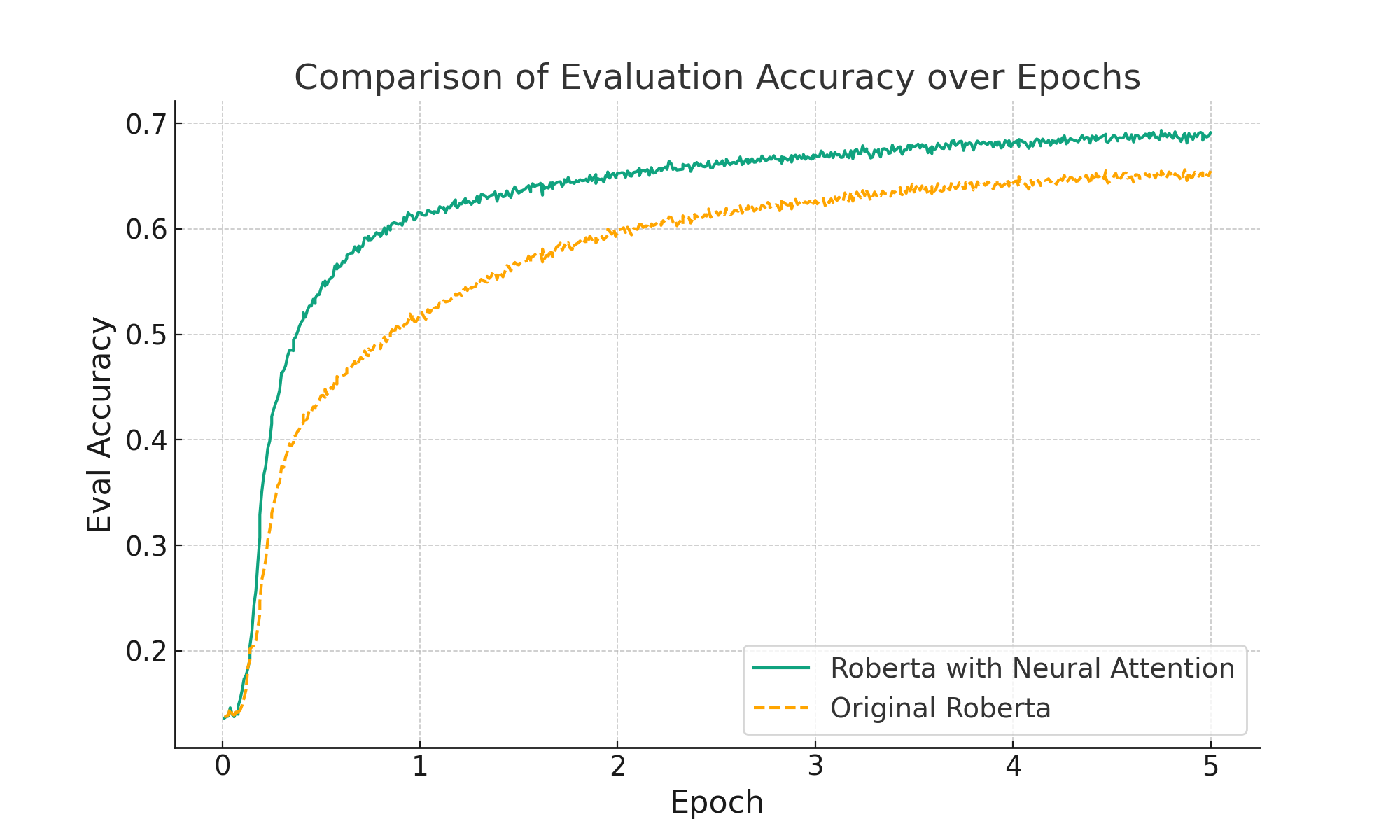}
\caption{Accuracy Comparison: Roberta with Neural Attention vs. Original Over Epochs}
\label{fig:acc_epoch_roberta}
\end{figure}

\begin{figure}[htbp]
\centering
\includegraphics[width=0.7\textwidth]{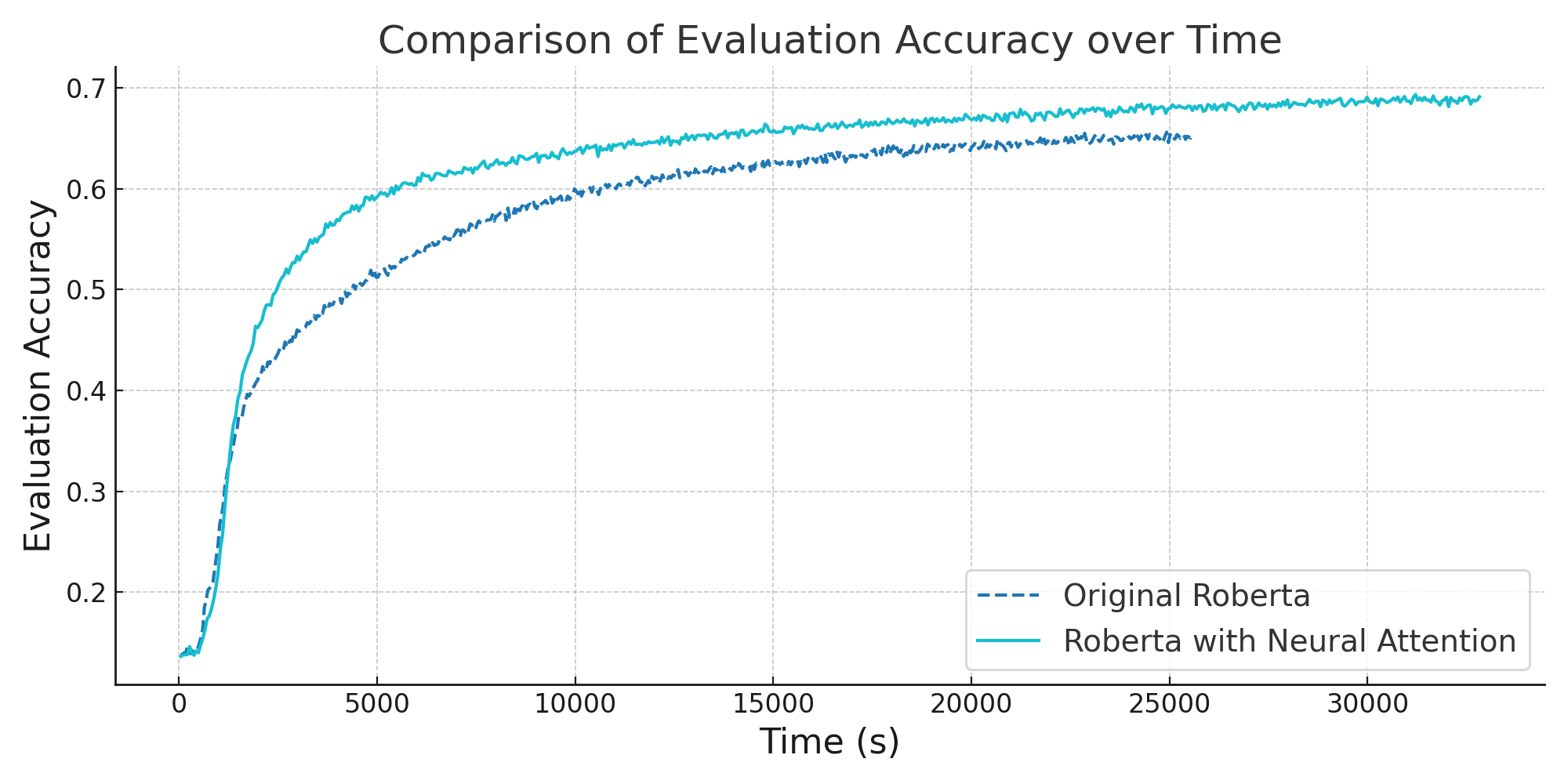}
\caption{Accuracy Comparison: Roberta with Neural Attention vs. Original Over Time (seconds)}
\label{fig:acc_time_roberta}
\end{figure}

Experimental results indicate that, compared to the original Roberta model, the modified model achieved a significant improvement in perplexity evaluation. The original model’s perplexity was 5.51, while the modified model was 4.47. Moreover, the modified model displayed a quicker improvement speed and higher final evaluation accuracy within the same training time. Figures \ref{fig:acc_epoch_roberta} and  \ref{fig:acc_time_roberta} show the accuracy graphs. It can be observed that the modified model’s accuracy improvement speed are faster than the original model. A detailed comparison of the perplexity and evaluation accuracy between the original and the modified model is presented in Table \ref{tab:comparison_roberta}.

\begin{table}[h]
\centering
\begin{tabular}{llll}
\toprule
Model   & Attention Mechanism Type & Perplexity & Eval Accuracy \\
\midrule
Roberta & Standard                 & 5.51 & 0.651\\
Roberta & Neural Attention         & 4.47 & 0.686\\
\bottomrule
\end{tabular}
\caption{Comparison between the original model and two different attention mechanism types in perplexity and eval\_accuracy.}
\label{tab:comparison_roberta}
\end{table}

\subsection{Experiment Two: Marian Model on Machine Translation Task}

\subsubsection{Model Modification and Training Setup}

We likewise modified the self-attention mechanism in the Marian model. The model was trained and evaluated on the iwslt2017-de-en dataset, undergoing a total of 6 training epochs.

\subsubsection{Results}

\begin{figure}[htbp]
\centering
\includegraphics[width=0.7\textwidth]{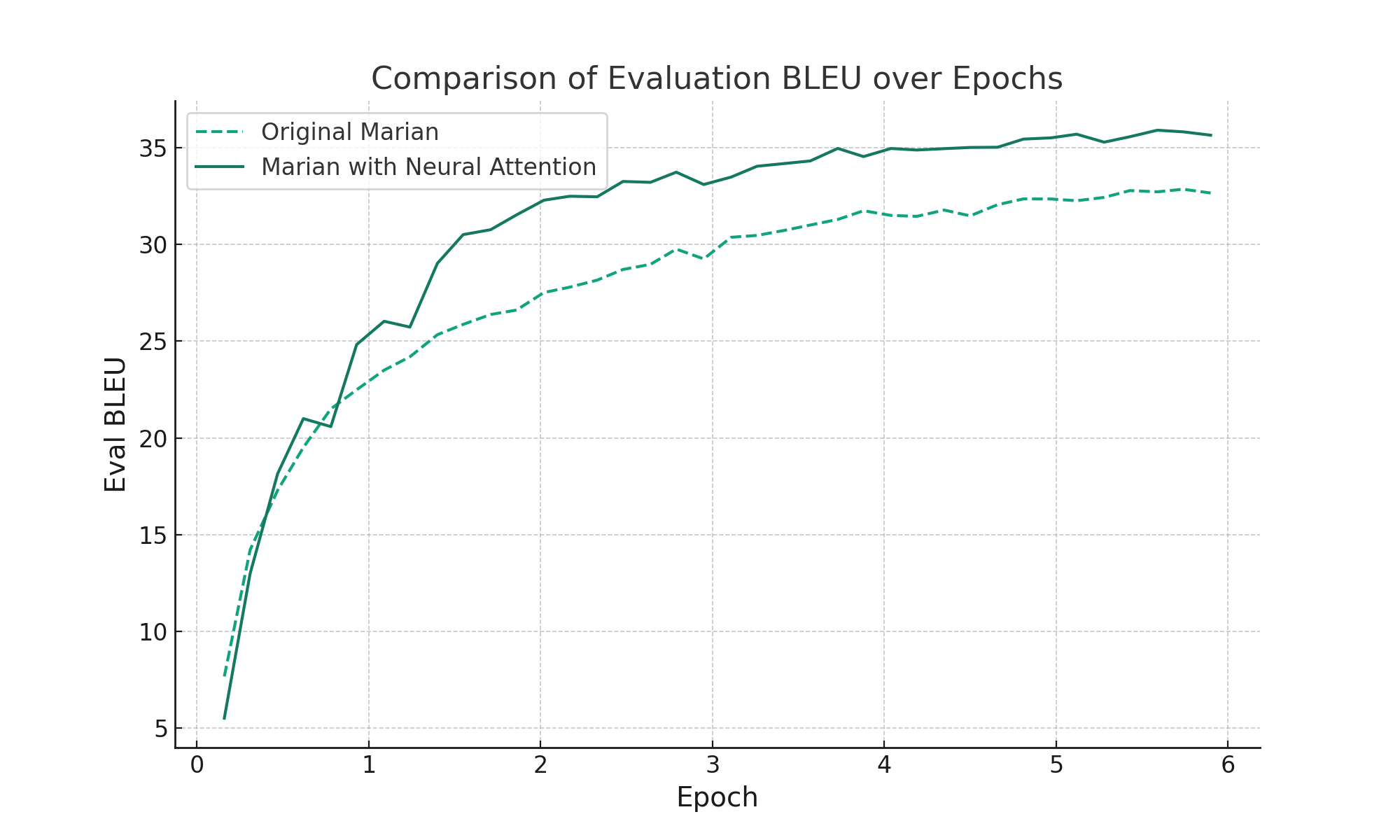}
\caption{BLEU Comparison: Marian with Neural Attention vs. Original Over Epochs}
\label{fig:bleu_epoch_marian}
\end{figure}

\begin{figure}[htbp]
\centering
\includegraphics[width=0.7\textwidth]{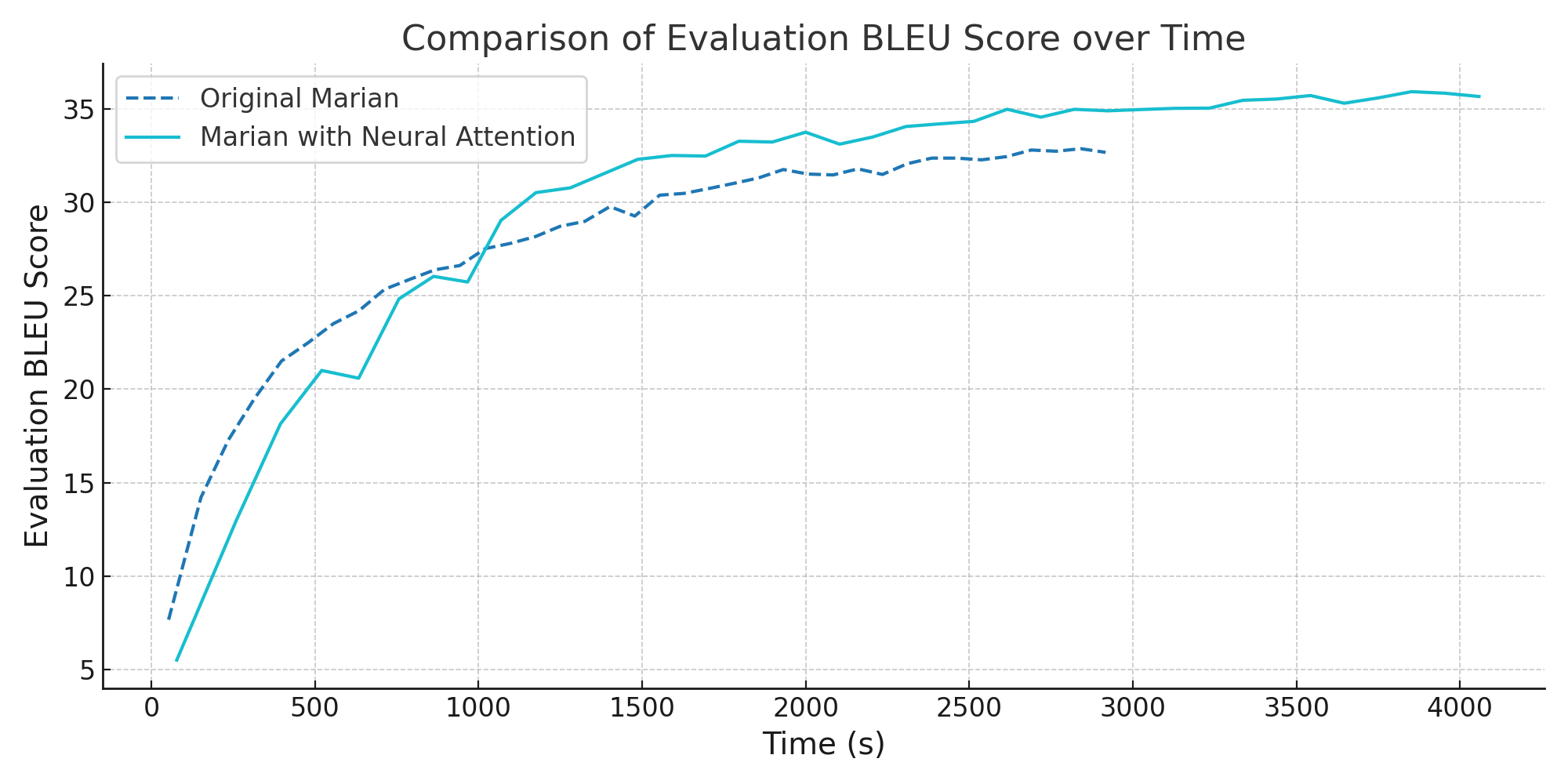}
\caption{BLEU Comparison: Marian with Neural Attention vs. Original Over Time (seconds)}
\label{fig:bleu_time_marian}
\end{figure}

Experimental results indicate that, compared to the original Marian model, the modified model achieved a significant improvement in the BLEU score. The original model’s BLEU score was 32.62, while the modified model was 35.76. As shown in Table \ref{tab:bleu_comparison}, both BLEU scores are compared. Figures \ref{fig:bleu_epoch_marian} show the BLEU score graphs over epochs, while Figures \ref{fig:bleu_time_marian} present them over time. From these graphs, it can similarly be observed that the modified model’s BLEU score improvement speed are faster than the original model.

\begin{table}[h]
\centering
\begin{tabular}{lll}
\toprule
Model  & Attention Mechanism Type & DE-EN BLEU \\
\midrule
Marian & Standard                 & 32.63       \\
Marian & Neural Attention         & 35.76       \\
\bottomrule
\end{tabular}
\caption{Comparison between the original model and two different attention mechanism types in the DE-EN BLEU score.}
\label{tab:bleu_comparison}
\end{table}

\subsection{Ablation Study}

In order to validate the pivotal role of the ReLU activation function in the enhancement of model performance, we conducted a comparison of methods for calculating keys (Key) and values (Value) in the Marian model on the iwslt2017-de-en dataset.

\textbf{Dual Linear Projection (DLP)}: A two-step linear mapping was utilized to compute the keys and values. Initially, the input is mapped to a twice-as-large intermediate representation through a linear layer, then it is mapped back to the original dimension through a second linear layer. Formally, given an input \(x \in \mathbb{R}^{d}\), the computation of keys and values is as follows:

\[
k, v = W_2 \cdot \text{LayerNorm}(W_1 \cdot x + b_1) + b_2
\]

\textbf{Neural Attention}: On top of DLP, we introduced a ReLU non-linear activation function between the two linear mappings. Therefore, the computation of keys and values becomes:

\[
k, v = W_2 \cdot \text{ReLU}(\text{LayerNorm}(W_1 \cdot x + b_1)) + b_2
\]

The results of the ablation study using these two mechanisms, along with the standard self-attention for comparison, are presented in Table 3.

\begin{table}[h]
\centering
\begin{tabular}{lll}
\toprule
Model  & Attention Mechanism Type   & DE-EN BLEU \\
\midrule
Marian & Standard Self-Attention    & 32.63       \\
Marian & Dual Linear Projection     & 32.64       \\
Marian & Neural Attention           & 35.76       \\
\bottomrule
\end{tabular}
\caption{BLEU score results on the iwslt2017-de-en dataset.}
\end{table}

Through the results presented in Table 3, we can observe that the non-linear layer (ReLU) plays a crucial role in the Neural Attention mechanism. The DLP method, which removes the ReLU activation function, did not demonstrate significant performance improvement, while the Neural Attention method markedly increased the BLEU score. This validates our initial intention and the necessity of introducing non-linear transformations.

\section{Conclusion}

This research delves deeply into the effectiveness and potential of employing Multi-Layer Perceptrons (MLP) for Query, Key, and Value (QKV) computation in the self-attention mechanism. Through a series of experiments and analyses on different models and tasks, we observed significant improvements in model performance in various aspects by introducing non-linear activation functions and complex key and value generation strategies.

In the masked language modeling task, we found that substituting the traditional linear transformation with an MLP neural network significantly improved the Roberta model in terms of Perplexity and other evaluation metrics. Similarly, in the machine translation task, the Marian model also showed a significant enhancement in the BLEU score after introducing a new attention mechanism computation method.

It is noteworthy that, due to computational resource limitations, we were unable to train a large model from scratch to attempt to surpass the existing state-of-the-art (SOTA) models. Our experiments mainly focused on existing pre-trained models and explored the feasibility and effects of fine-tuning models by modifying the self-attention mechanism.

Although we have made some positive strides in this research, there is still ample space for exploration and optimization in using an MLP neural network as the QKV computation method in the self-attention mechanism. Future research directions could include exploring different network architectures, activation functions, and the potential applicability of this method on other NLP tasks and models.

In summary, this research offers a novel approach and perspective, showcasing the potential value of introducing non-linearity and complex computations into the self-attention mechanism, and lays the foundation for further research and exploration. We hope these findings provide insights for researchers and developers in the field of natural language processing and further propel the development and innovation of attention mechanisms.

\bibliographystyle{unsrtnat}
\bibliography{neural_networks_paper.tex}  

\begin{thebibliography}{12}
\providecommand{\natexlab}[1]{#1}
\providecommand{\url}[1]{\texttt{#1}}
\expandafter\ifx\csname urlstyle\endcsname\relax
  \providecommand{\doi}[1]{doi: #1}\else
  \providecommand{\doi}{doi: \begingroup \urlstyle{rm}\Url}\fi

\bibitem[Vaswani et~al.(2017)Vaswani, Shazeer, Parmar, Uszkoreit, Jones, Gomez, Kaiser, and Polosukhin]{vaswani2017attention}
Ashish Vaswani, Noam Shazeer, Niki Parmar, Jakob Uszkoreit, Llion Jones, Aidan~N Gomez, {\L}ukasz Kaiser, and Illia Polosukhin.
\newblock Attention is all you need.
\newblock \emph{Advances in neural information processing systems}, 30, 2017.

\bibitem[Goodfellow et~al.(2016)Goodfellow, Bengio, and Courville]{goodfellow2016deep}
Ian Goodfellow, Yoshua Bengio, and Aaron Courville.
\newblock \emph{Deep learning}.
\newblock MIT press, 2016.

\bibitem[Dosovitskiy et~al.(2020)Dosovitskiy, Beyer, Kolesnikov, Weissenborn, Zhai, Unterthiner, Dehghani, Minderer, Heigold, Gelly, et~al.]{dosovitskiy2020image}
Alexey Dosovitskiy, Lucas Beyer, Alexander Kolesnikov, Dirk Weissenborn, Xiaohua Zhai, Thomas Unterthiner, Mostafa Dehghani, Matthias Minderer, Georg Heigold, Sylvain Gelly, et~al.
\newblock An image is worth 16x16 words: Transformers for image recognition at scale.
\newblock \emph{arXiv preprint arXiv:2010.11929}, 2020.

\bibitem[Lin et~al.(2017)Lin, Feng, Santos, Yu, Xiang, Zhou, and Bengio]{lin2017structured}
Zhouhan Lin, Minwei Feng, Cicero Nogueira~dos Santos, Mo~Yu, Bing Xiang, Bowen Zhou, and Yoshua Bengio.
\newblock A structured self-attentive sentence embedding.
\newblock \emph{arXiv preprint arXiv:1703.03130}, 2017.

\bibitem[Liu et~al.(2019)Liu, Ott, Goyal, Du, Joshi, Chen, Levy, Lewis, Zettlemoyer, and Stoyanov]{liu2019roberta}
Yinhan Liu, Myle Ott, Naman Goyal, Jingfei Du, Mandar Joshi, Danqi Chen, Omer Levy, Mike Lewis, Luke Zettlemoyer, and Veselin Stoyanov.
\newblock Roberta: A robustly optimized bert pretraining approach.
\newblock \emph{arXiv preprint arXiv:1907.11692}, 2019.

\bibitem[Devlin et~al.(2018)Devlin, Chang, Lee, and Toutanova]{devlin2018bert}
Jacob Devlin, Ming-Wei Chang, Kenton Lee, and Kristina Toutanova.
\newblock Bert: Pre-training of deep bidirectional transformers for language understanding.
\newblock \emph{arXiv preprint arXiv:1810.04805}, 2018.

\bibitem[Junczys-Dowmunt et~al.(2018)Junczys-Dowmunt, Grundkiewicz, Dwojak, Hoang, Heafield, Neckermann, Seide, Germann, Aji, Bogoychev, et~al.]{junczys2018marian}
Marcin Junczys-Dowmunt, Roman Grundkiewicz, Tomasz Dwojak, Hieu Hoang, Kenneth Heafield, Tom Neckermann, Frank Seide, Ulrich Germann, Alham~Fikri Aji, Nikolay Bogoychev, et~al.
\newblock Marian: Fast neural machine translation in c++.
\newblock \emph{arXiv preprint arXiv:1804.00344}, 2018.

\bibitem[Rumelhart et~al.(1986)Rumelhart, Hinton, and Williams]{rumelhart1986learning}
David~E Rumelhart, Geoffrey~E Hinton, and Ronald~J Williams.
\newblock Learning representations by back-propagating errors.
\newblock \emph{nature}, 323\penalty0 (6088):\penalty0 533--536, 1986.

\bibitem[Nair and Hinton(2010)]{nair2010rectified}
Vinod Nair and Geoffrey~E Hinton.
\newblock Rectified linear units improve restricted boltzmann machines.
\newblock In \emph{Proceedings of the 27th international conference on machine learning (ICML-10)}, pages 807--814, 2010.

\bibitem[Ba et~al.(2016)Ba, Kiros, and Hinton]{ba2016layer}
Jimmy~Lei Ba, Jamie~Ryan Kiros, and Geoffrey~E Hinton.
\newblock Layer normalization.
\newblock \emph{arXiv preprint arXiv:1607.06450}, 2016.

\bibitem[Wolf et~al.(2020)Wolf, Debut, Sanh, Chaumond, Delangue, Moi, Cistac, Rault, Louf, Funtowicz, Davison, Shleifer, von Platen, Ma, Jernite, Plu, Xu, Scao, Gugger, Drame, Lhoest, and Rush]{wolf-etal-2020-transformers}
Thomas Wolf, Lysandre Debut, Victor Sanh, Julien Chaumond, Clement Delangue, Anthony Moi, Pierric Cistac, Tim Rault, Rémi Louf, Morgan Funtowicz, Joe Davison, Sam Shleifer, Patrick von Platen, Clara Ma, Yacine Jernite, Julien Plu, Canwen Xu, Teven~Le Scao, Sylvain Gugger, Mariama Drame, Quentin Lhoest, and Alexander~M. Rush.
\newblock Transformers: State-of-the-art natural language processing.
\newblock In \emph{Proceedings of the 2020 Conference on Empirical Methods in Natural Language Processing: System Demonstrations}, pages 38--45, Online, October 2020. Association for Computational Linguistics.
\newblock URL \url{https://www.aclweb.org/anthology/2020.emnlp-demos.6}.

\bibitem[Tiedemann and Thottingal(2020)]{TiedemannThottingal:EAMT2020}
J{\"o}rg Tiedemann and Santhosh Thottingal.
\newblock {OPUS-MT} — {B}uilding open translation services for the {W}orld.
\newblock In \emph{Proceedings of the 22nd Annual Conferenec of the European Association for Machine Translation (EAMT)}, Lisbon, Portugal, 2020.

\end{thebibliography}

\end{document}